\documentclass[letterpaper]{article} % DO NOT CHANGE THIS
\usepackage{aaai2026}  % DO NOT CHANGE THIS
\usepackage{times}  % DO NOT CHANGE THIS
\usepackage{helvet}  % DO NOT CHANGE THIS
\usepackage{courier}  % DO NOT CHANGE THIS
\usepackage[hyphens]{url}  % DO NOT CHANGE THIS
\usepackage{graphicx} % DO NOT CHANGE THIS
\usepackage{natbib}  % DO NOT CHANGE THIS AND DO NOT ADD ANY OPTIONS TO IT
\usepackage{caption} % DO NOT CHANGE THIS AND DO NOT ADD ANY OPTIONS TO IT
\usepackage{amsmath} 
\usepackage{amsfonts}
\usepackage{algorithm}
\usepackage{algorithmic}

\usepackage{newfloat}
\usepackage{listings}
\DeclareCaptionStyle{ruled}{labelfont=normalfont,labelsep=colon,strut=off} % DO NOT CHANGE THIS
\floatstyle{ruled}
\newfloat{listing}{tb}{lst}{}
\floatname{listing}{Listing}
\title{DCMM-Transformer: Degree-Corrected Mixed-Membership Attention for Medical Imaging}
\author{
    Huimin Cheng\textsuperscript{\rm 1}\equalcontrib, Xiaowei Yu\textsuperscript{\rm 2}\equalcontrib, Shushan Wu\textsuperscript{\rm 3}, Luyang Fang\textsuperscript{\rm 3}, Chao Cao\textsuperscript{\rm 4}, \\
    Jing Zhang\textsuperscript{\rm 4}, Tianming Liu\textsuperscript{\rm 5}, Dajiang Zhu\textsuperscript{\rm 4}, Wenxuan Zhong\textsuperscript{\rm 3}, Ping Ma\textsuperscript{\rm 3}
    \\
}
\affiliations{
    \textsuperscript{\rm 1}Department of Biostatistics, Boston University \\
    \textsuperscript{\rm 2}Department of Computer Science, Missouri University of Science and Technology \\
    \textsuperscript{\rm 3}Department of Statistics, University of Georgia \\
    \textsuperscript{\rm 4}Department of Computer Science and Engineering, University of Texas at Arlington, USA \\
    \textsuperscript{\rm 5}School of Computing,  University of Georgia, USA 
    
}

\usepackage{bibentry}
\begin{document}

\maketitle

\begin{abstract}
Medical images exhibit latent anatomical groupings, such as organs, tissues, and pathological regions, that standard Vision Transformers (ViTs) fail to exploit.
While recent work like SBM-Transformer attempts to incorporate such structures through stochastic binary masking, they suffer from non-differentiability, training instability, and the inability to model complex community structure. 
We present DCMM-Transformer, a novel ViT architecture for medical image analysis that incorporates a Degree-Corrected Mixed-Membership (DCMM) model as an additive bias in self-attention. Unlike prior approaches that rely on multiplicative masking and binary sampling, our method introduces community structure and degree heterogeneity in a fully differentiable and interpretable manner. 
Comprehensive experiments across diverse medical imaging datasets, including brain, chest, breast, and ocular modalities, demonstrate the superior performance and generalizability of the proposed approach. Furthermore, the learned group structure and structured attention modulation substantially enhance interpretability by yielding attention maps that are anatomically meaningful and semantically coherent.

\end{abstract}


\section{Introduction}

% Medical image analysis is important. 
% Recent development in VIT has successful applications in medical images. 
% An important feature of medical image is it contain inherent structural organization... 
% Standard Transformer treat equally... 

Medical image analysis is essential for disease diagnosis, prognosis, and informing clinical decisions \citep{varoquaux2022machine, yu2023supervised, yu2024cp, yu2024core}.
Vision Transformers (ViTs) and their variants, such as TransUNet \citep{chen2024transunet} and Swin Transformer \citep{liu2021swin}, have achieved state-of-the-art performance.
These models segment images into small, fixed-size patches (tokens) and employ self-attention mechanisms to model complex, long-range relationships across different anatomical structures. 
By capturing such dependencies, ViTs have demonstrated impressive results across a range of tasks, including image classification, segmentation, and object detection \citep{ hatamizadeh2022unetr, yu2023noisynn}.

%  Its key strength stems from the multi-head attention module, where
% a so-called attention score matrix computes how contextually important one token is to another for all
% possible token pairs.

\textbf{Challenges for Medical Images. }
Despite rapid advances, medical image analysis remains challenging and often necessitates architectural designs tailored to the unique properties of medical data \citep{huang2025semi, huang2024interlude, yu2022long, litjens2017survey, yu2022disentangling, zhang2025brain, yu2025domain}.
Medical images are spatially organized into distinct anatomical regions, such as tumor areas, healthy tissue, and blood vessels, that form meaningful clusters within the image \citep{esteva2021deep, yu2021free, chavoshnejad2023an}. 
% For example, a histopathology slide used in cancer diagnostics typically contains distinct regions corresponding to tumor tissue and healthy tissue \citep{gurcan2009histopathological}.
Patches from the same anatomical or pathological region tend to be correlated, sharing features that are particularly informative for clinical tasks. Conversely, patches from different regions (such as tumor versus healthy tissue, or across distinct organs) often exhibit distinct underlying biological properties, resulting in weaker relationships between them. 

\textbf{Motivation for Community-Aware Attention.}
This community characteristic motivates the development of community-aware attention mechanisms, which encourage the model to focus on interactions within anatomically or pathologically coherent regions \citep{wang2019weakly}. Such approaches are particularly valuable for detecting subtle patterns: in tumor detection, for example, the signal from any single patch within a tumor may be weak or ambiguous. By aggregating information across patches within the same community, the model can amplify these weak signals, resulting in a more coherent and accurate diagnostic assessment \citep{ilse2018attention, yu2023core}.

\textbf{Limitations of Standard ViTs.}
Nevertheless, conventional transformer-based attention mechanisms do not take this underlying community structure into account. Traditional attention approaches treat tokens uniformly, calculating weights solely based on learned feature similarities without considering anatomical groupings \citep{vaswani2017attention, dosovitskiy2021image}. As a result, the model may sometimes focus on regions that are distant or unrelated in a clinical sense, simply because they appear similar in feature space. This lack of anatomical awareness can cause the model to overlook clinically important regions, become more vulnerable to noise or artifacts, and make its predictions harder to interpret  \citep{li2023vision}. These limitations underscore the necessity for transformers that explicitly incorporate community structure \citep{li2023transforming}.

\textbf{Limitations of SBM-Transformer.}
Recent work has explored incorporating explicit community structures into attention mechanisms. One notable example is SBM-Transformer \citep{cho2022transformers} which integrates a mixed-membership Stochastic Block Model (SBM) into each attention head to learn latent communities among image patches. To reduce computational costs, the SBM-Transformer employs binary attention masks that randomly sample token connections based on probabilities learned from the SBM. Only token pairs selected by these masks can attend to each other, resulting in sparse attention. However, this binary masking introduces non-differentiability, which requires surrogate gradient methods like the Straight-Through Estimator and leads to optimization bias and convergence instability \citep{cho2022transformers} . Additionally, collapsing probabilistic connection strengths into hard, binary decisions discards useful information about the strength of token relationships and increases variance during both training and inference.
A further limitation is that the SBM-Transformer ignores degree heterogeneity. In practice, image patches exhibit significant variation in connectivity and importance. For example, tumor cores often act as local hubs by interacting with many other regions, an effect not captured under the assumption of uniform connectivity.

% (1) Overlapping Community Structure: An image batch may contain part of a tumor, healthy tissue.  In such cases, a patch's identity is best represented as a mixture of multiple communities, rather than a hard assignment. 

\def \our{\mbox{DCMM-Transformer}}

\begin{figure}[!htb]
\centering
\includegraphics[width=1.0\columnwidth]{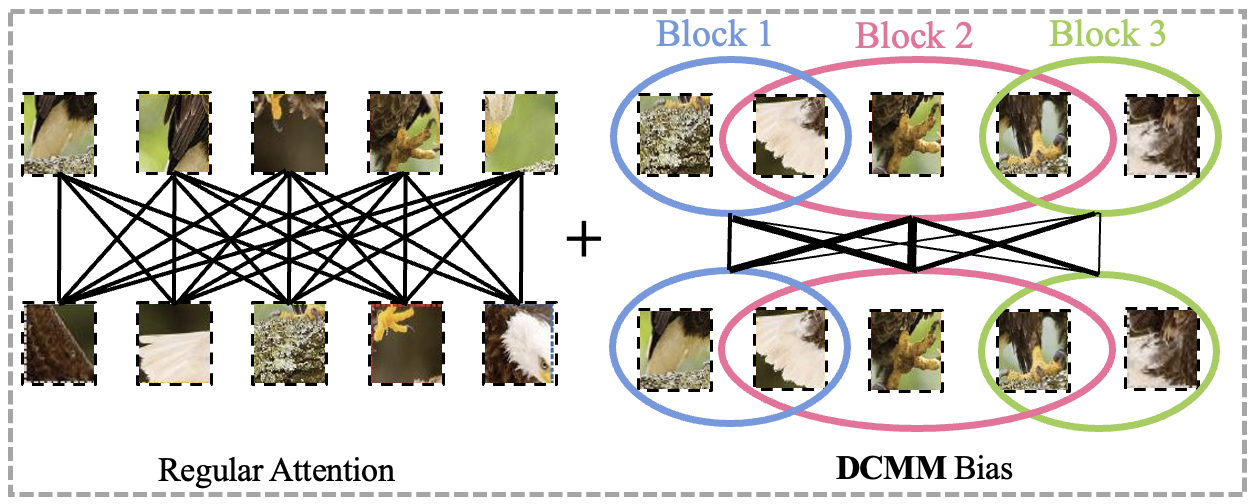} % Reduce the figure size so that it is slightly narrower than the column. Don't use precise values for figure width.This setup will avoid overfull boxes.
\vspace{-20pt}
\caption{Overview of the $\our$. The DCMM model imposes latent community structure among image patches, grouping them into blocks (e.g., Blocks 1–3 in blue, red, and green), and assigns probabilities to the interactions within and between these communities. This community-based bias is then added to the standard attention logits.}
\label{DCMMAtt}
\end{figure}

\textbf{Proposed Method.}
To overcome these limitations, we propose $\our$, 
a method integrating a Degree-Corrected Mixed-Membership (DCMM) model into ViTs. An overview of $\our$ is shown in Figure~\ref{DCMMAtt}.
In particular, we model the latent community structure among image patches using the DCMM framework, which captures both overlapping community membership and degree heterogeneity. The DCMM module generates a learned probability matrix, where each entry reflects the likelihood of interaction between a pair of patches, informed by their shared community memberships and individual connectivity.
Our approach adds the learned probability matrix directly to attention logits as an additive bias, bypassing random sampling and binary discretization. 
This community-aware bias helps the model aggregate weak or distributed signals from related patches, such as those within the same tumor region, thereby enhancing the detection of clinically important patterns that might be missed when treating all patches equally.

\textbf{Advantages.}
DCMM-Transformer offers three key advantages: (1) Structural guidance without distortion: Unlike the SBM-Transformer \citep{cho2022transformers}, which uses multiplicative masking to impose sparsity and risks distorting semantic relationships by over-suppressing weak connections or unevenly amplifying others, our additive bias offers structural guidance while preserving data-driven insights. (2) Training stability: Addition maintains stable gradients and normalization, avoiding vanishing gradient issues. (3) Enhanced modeling: DCMM captures both degree heterogeneity and mixed membership, capabilities absent in standard SBMs.
We provide extensive evaluations across five medical image classification tasks, demonstrating an average $3.7\%$ improvement in performance over standard ViT and its variants, as well as enhanced interpretability for clinical applications. Ablation studies further reveal that excluding degree heterogeneity reduces performance by $2.3\%$.

\section{Related Work}

In this section, we present two areas of related work:  SBMs and Vision Transformers.

\subsection{Related Work on SBM}

\textbf{SBM.}
The  stochastic block model \cite{holland1983stochastic, cheng2018community, liu2019evaluation} is a fundamental probabilistic model for networks with community structure. In the SBM, each node is assigned to one of $L$ latent communities (or blocks), and the probability of an edge between any two nodes depends only on their community memberships. Let $A \in \left(0,1\right)^{n \times n}$ denote the adjacency matrix of the network, and let $z_i \in {1,\dots,L}$ be the community label of node $i$. The block matrix $B \in (0,1)^{L \times L}$ specifies the probability of connection between communities, so that edges are sampled independently according to
\begin{equation}
  \Pr(A_{ij}=1)=B_{z_i z_j},
  \label{eq:sbm}
\end{equation}
where $B_{z_i, z_j}$ is the probability of an edge between node $i$ and node $j$.
While the SBM provides a simple and effective framework for modeling community structure, it assumes that all nodes within a block have identical expected degrees, which is often violated in real-world networks exhibiting heterogeneous degree distributions.

\textbf{Degree-Corrected SBM.}
To address the limitation of the homogeneous degree assumption in SBM, the Degree-Corrected Stochastic Block Model (DC-SBM)~\cite{karrer2011stochastic} introduces node-specific degree parameters. Under DC-SBM, the probability of an edge between nodes $i$ and $j$ depends not only on their community memberships, but also on individual propensity parameters $\theta_i>0$ and $\theta_j>0$ that control expected node degrees, allowing for more realistic modeling of networks with heterogeneous degrees:
\begin{equation}
  \Pr(A_{ij}=1)=\theta_i\,\theta_j\,B_{z_i z_j}.
  \qquad
  % \mathbb E[\theta_i]=1
  % \text{ for identifiability.}
  \label{eq:dcsbm}
\end{equation}

\textbf{{Mixed‑Membership SBM}}.
The Mixed Membership SBM (MMSBM) \citep{airoldi2008mixed} extends the classical SBM to handle overlapping communities by replacing hard community assignments with soft membership vectors.  Each
node $i$ has a probability membership vector:  
$\boldsymbol{\pi}_i=(\pi_i(1),\dots,\pi_i(L))^{\!\top}\!\in\! [0,1]^{L}$, where $\sum_{\ell=1}^{L}\pi_i(\ell)=1.$
The probability of an edge between nodes $i$ and $j$ is
\begin{equation}
  \Pr(A_{ij}=1)=
  \sum_{\ell=1}^{L}\sum_{\ell'=1}^{L}
     \pi_i(\ell)\,\pi_j(\ell')\,B_{\ell\ell'}
  =\boldsymbol{\pi}_i^{\!\top}B\boldsymbol{\pi}_j.
  \label{eq:mmsbm}
\end{equation}
When each $\boldsymbol{\pi}_i$ is a standard basis vector (i.e., nodes belong to only one community), Equation \eqref{eq:mmsbm} reduces to \eqref{eq:sbm} in classical SBM.

\textbf{{Degree‑Corrected Mixed‑Membership SBM (DCMM)}}.
The most general model combines mixed membership and  degree heterogeneity \citep{jin2024mixed}:
\begin{equation}
  \Pr(A_{ij}=1)=
    \theta_i\,\theta_j\,
    \boldsymbol{\pi}_i^{\!\top}B\boldsymbol{\pi}_j.
  \label{eq:dcmmsbm}
\end{equation}
Previous models are special cases of DCMM: setting $\theta_i = 1$ recovers the MMSBM, while taking both $\theta_i = 1$ and one-hot $\boldsymbol{\pi}_i$ yields the classical SBM.

\subsection{Related Work on Transformer}

\textbf{Vision Transformer and Attention.} 
% ViT has emerged as a powerful architecture for a variety of computer vision tasks, including medical image analysis.
ViTs process images by treating them as sequences of tokens, similar to how language models handle text.  First, the input image is divided into $n$ non-overlapping, flattened 2D patches (tokens), resulting in a matrix $X \in \mathbb{R}^{n \times d}$, where $d$ is the dimension of each patch embedding. The core of ViT is the self-attention mechanism, which enables the model to capture dependencies and interactions between all pairs of patches.
% Its core component is the scaled dot-product self-attention mechanism.
% For an input image, which is first divided into a sequence of flattened 2D patches (tokens) $X \in \mathbb{R}^{n \times d}$,
Second, within each attention head, the model computes pairwise relationships among tokens by projecting them into query ($Q$), key ($K$), and value ($V$) representations using learnable weight matrices
$Q = XW^Q, \quad K = XW^K, \quad V = XW^V,$
where $W^Q, W^K, W^V \in \mathbb{R}^{d \times d_h}$ and $d_h$ is the head dimension.
 Attention scores are then computed as the scaled dot product $\frac{QK^\top}{\sqrt{d_h}}$, followed by row-wise softmax normalization to obtain weights $\sigma\left(\frac{QK^\top}{\sqrt{d_h}}\right)$ by softmax. The output is finally derived as a weighted sum of values: 
 $$\text{Attn}(Q, K, V) = \sigma\left(\frac{QK^\top}{\sqrt{d_h}}\right) V.$$ This process allows each patch to attend to all other patches based on their learned similarities, enabling the model to capture long-range dependencies across the entire image.

\textbf{SBM-Transformer.}
SBM-Transformer \cite{cho2022transformers}  directly replaces the dense attention mechanism with a sparse, learnable alternative guided by an MMSBM, with the main purpose of reducing the computational burden of the Transformer. In particular, the connection probability $p_{ij}$ between tokens $i$ and $j$ is given by Equation \eqref{eq:mmsbm}, where $\mathbf{\pi}$ and $\mathbf{B}$ are parameters learned from neural networks. 
A binary attention mask $M_{ij}$ is then sampled as $M_{ij} \sim \text{Bernoulli}(p_{ij})$.
This mask is applied to the attention mechanism via masked attention:
$$\text{Attn}_{\text{mask}}(Q, K, V, M) = \sigma_M \left( M \odot \frac{Q K^\top}{\sqrt{d_h}} \right) V,$$ where $\odot$ denotes element-wise multiplication, and $\sigma_M(\cdot)$ is the masked softmax that normalizes only over the non-zero entries in each row of $M$. 

Even though SBM-Transformer reduces computational cost, it suffers from these limitations: (1) Non-differentiability and optimization instability: The binary sampling is non-differentiable,  as the sampling function lacks a continuous derivative with respect to $ p_{ij} $. This breaks gradient flow, requiring surrogate methods like Straight-Through Estimator, but this surrogate introduces bias and high variance.  (2) Information loss through hard thresholding: Binarization discards the probabilistic information in $p_{ij}$, which can encode subtle relationships. For example, $p_{ij} = 0.99$ and $p_{ij} = 0.1$ might both yield $M_{ij} = 1$, but their connection strengths differ significantly. (3) Lack of degree heterogeneity in the MMSBM: MMSBM fails to capture degree heterogeneity, a characteristic of real networks, including image patch graphs where hub patches (e.g., tumor cores) connect broadly, while peripherals do not. 

% The core idea is to treat the queries and keys within an attention head as nodes in a bipartite graph. Instead of computing the affinity between all query-key pairs, the model first learns a community structure for the tokens and then samples a sparse graph of connections based on the learned SBM parameters.
% CAS-Trans proposes a novel positional encoding aware of abstract syntax tree (AST) node context and SBM attention
%  for the code summarization task \citep{oh2024csa}.

 \textbf{Transformer with Additive Bias.}
Another related line of research enhances attention mechanisms with explicit bias terms that encode spatial or structural relationships. For example, ACC-ViT \citep{ibtehaz2024acc} introduces Atrous Attention with multiple dilation rates: $\text{Attn}(Q, K, V) = \operatorname{Softmax}\left(\frac{Q K^\top}{\sqrt{d}} + b\right) V$, where $b$ denotes branch-specific relative positional biases, and outputs are fused through gated aggregation. The Swin Transformer \citep{liu2021swin} restricts attention to local patch blocks while adding a learnable relative position bias for substantial gains. Similarly, MaxViT \citep{tu2022maxvit} employs multi-axis self-attention to outperform standard mechanisms. However, these methods do not take into account the community structure. 

In sum, while prior works have shown the promise of community-aware attention  and additive bias, they remain limited, e.g.,  SBM-based masking is non-differentiable, while standard bias terms do not consider the latent community structure.

\begin{figure*}[t]
\centering
\includegraphics[width=2.0\columnwidth]{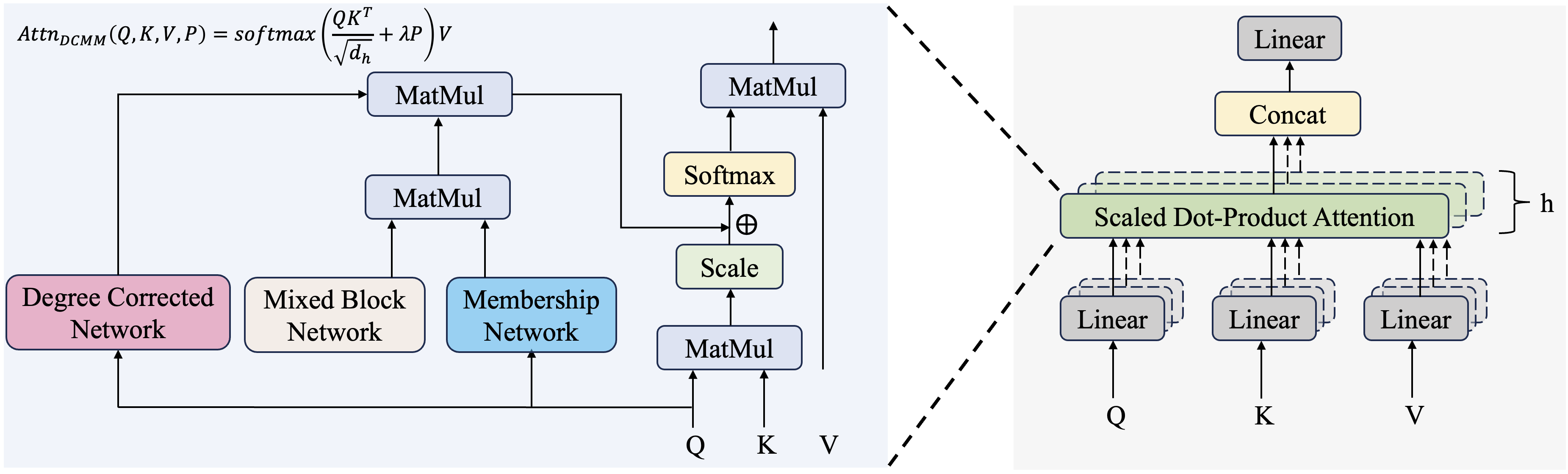} 
\vspace{-5pt}
\caption{Overview of the proposed DCMM-Transformer mechanism integrated into the self-attention module of Vision Transformers. For each token, a soft group membership vector and a degree scalar are constructed from the query representation. These are used to construct a structured attention bias based on learnable inter-group affinities. The bias is added to the standard attention scores to modulate them before softmax.}
\label{DCMMViT}
\end{figure*}

%Our Method:
\section{ $\our$}
To address these limitations, the DCMM-Transformer augments the standard transformer attention logits with a connection probability matrix under DCMM, allowing the model to learn both soft community assignments and node-specific centrality in a fully differentiable manner.

% it preserves differentiability, captures degree heterogeneity, and provides richer, more interpretable community modeling for image patch relationships.

\subsection{Overview of $\our$}
The key innovation of $\our$ is employing a novel self-attention logits by adding a learned community structure matrix $P \in \mathbb{R}^{n \times n}$ which is generated under the DCMM:
\begin{equation}\label{eq:our}
    \text{Attn}_{\text{DCMM}}(Q,K,V,P) = \sigma(\frac{QK^\top}{\sqrt{d_h}} + \lambda P)V,
\end{equation}
where 
$\lambda$ is a tunable hyperparameter controlling the contribution of the community structure.
When 
$\lambda = 0$, the model reduces to a standard transformer; as 
$\lambda$ increases, the community structure dominates. 

Figure~\ref{DCMMViT} illustrates the DCMM‑Transformer.
The DCMM bias $P$ is constructed on the left side of the diagram through a series of specialized networks and operations applied primarily to $Q$. Specifically, each token (image patch) $i$ is processed by two parallel neural networks: The membership network module computes soft community membership vectors $\mathbf{\pi}_i$ for each patch, indicating its association with each latent community. The degree corrected network module produces a node-specific degree parameter $\theta_i$, capturing the relative centrality or importance of each patch. The mixed block network module introduces $L$ trainable cluster embeddings for latent communities.

\textbf{Rationale of additive bias. } We choose additive bias over multiplicative bias in Equation \eqref{eq:our} due to the following reasons. (1)
\textbf{Gentle Enhancement vs. Exponential Scaling}. When the DCMM bias $P$ is added to the logits, the effect after the softmax is a multiplication by $\exp(\lambda P_{ij})$, where $P_{ij}$ is always between 0 and 1. This means the community prior can gently increase the corresponding attention weights by at most a factor of $e^{\lambda}$ (when $P_{ij}=1$), or leave them nearly unchanged (when $P_{ij}$ is close to 0), without erasing the information from the original feature similarities. In contrast, with multiplicative bias, the logits are directly scaled by $P_{ij}$ before exponentiation, so any value of $P_{ij}$ less than 1 can drastically shrink the effective logit, leading to a nearly uniform or uninformative attention pattern and causing the model to overlook meaningful relationships. 
(2) \textbf{Training Stability. } Let $l=\frac{QK^\top}{\sqrt{d_h}}$. For additive bias, the softmax gradient is: $\frac{\partial \sigma(l + \lambda P)}{\partial l} = \text{diag}(\sigma(l + \lambda P)) - \sigma(l + \lambda P) \sigma(l + \lambda P)^\top$. This standard softmax Jacobian remains well-conditioned regardless of bias values $\lambda P$, since the gradient structure is unchanged—only the softmax outputs shift. However, for multiplicative bias $P \odot l$, the gradient becomes: $\frac{\partial \sigma(P \odot l)}{\partial l} = \text{diag}(P ) \cdot [\text{diag}(\sigma(P \odot l)) - \sigma(P  \odot l) \sigma(P  \odot l)^\top]$
 Since the $\text{diag}(P)$ scales each gradient component by the corresponding $ P_{ij} \in [0,1] $, this can lead to  a vanishing gradient problem for low $P_{ij}$ which is very common in sparse networks. 
%(3) \textbf{Empirically}, our results in Table \ref{tab:ASintegration} in Appendix show that additive bias significantly outperforms multiplicative bias. 

% The detailed formulations of DCMM-Transformer and the associated training loss are detailed in the following. We first illustrate the forward step of our attention module and how the underlying DCMM  of each head is parameterized by the input tensors.  

\begin{table*}[!htb]
\centering
\caption{Comparative classification accuracy (\%) of our DCMM-Transformer and baseline Vision Transformer variants across five medical imaging datasets. Bold values indicate the highest accuracy for each dataset.}
\vspace{-5pt}
\begin{tabular}{ccccccc}
\hline
Method           & ChestXray  & SIIMACR & INbreast & ADNI & EyeDisease  & Average \\ \hline
ViT              & 92.3      & 86.9    & 79.7     & 66.7 & 92.5    & 83.6  \\
SBM-Transformer     & 95.8     & 85.8    & 78.4    & 64.2 & 91.3   & 83.1   \\ 
DeiT             & 95.8      & 86.8    & 83.8     & 65.4 & 91.7     & 84.5  \\
ConViT           & 93.6      & 79.6    & 81.1     & 71.6 & 92.4    & 83.7   \\
Swin Transformer & 95.0      & 87.6    & 84.8     & 65.4 & 92.9     & 85.1  \\
MaxViT           & 94.5      & 86.4    & 83.3     & 71.7 & 93.2    & 85.8   \\ 
\hline
\textbf{DCMM-Transformer} & \textbf{96.0}      & \textbf{88.2}    & \textbf{85.1}     & \textbf{74.1} & \textbf{93.3}   & \textbf{87.3}   \\ \hline
\end{tabular}
\label{tab:method_comparison}
\end{table*}

\subsection{Procedures of $\our$}

We detail below how $\our$ constructs and updates the community structure matrix $P$ within the model pipeline.

\textbf{(1) Cluster Embedding Initialization:}
To represent latent communities, we introduce $L$ trainable cluster embeddings $C \in \mathbb{R}^{L \times d_l}$.  These cluster embeddings parameterize the features of each community within the attention head’s representation space.

\textbf{(2) Block Connecting Probability.}
To model the probability of interaction between communities, we construct a non-negative block connection probability matrix $B \in (0,1)^{L \times L}$. Each entry $B_{ij}$ represents the probability of connection between community $i$ and community $j$. The diagonal entries of $B$ indicate intra-community connection probabilities, while the off-diagonal entries represent inter-community probabilities. We obtain $B$ by applying a row-wise softmax to the inner product $C C^\top$, where $C$ is the community embeddings. 

% A non-negative block connecting matrix ${B} \in (0,1)^{K \times K}$, is constructed by applying a row-wise softmax to the inner product of the cluster embeddings, i.e., ${B} = \text{softmax}(C C^\top)$. This matrix quantifies the interaction strengths between different communities, allowing for flexible, data-driven modeling of intra- and inter-group relationships.

\textbf{(3) Soft Community Memberships:}
For each token $i$, we then construct its soft community membership vector $\mathbf{\pi}_i \in [0,1]^L$, where each element $\mathbf{\pi}_{ij}$ represents how strongly token $i$ belongs to community $j$. 
Specifically,   $\mathbf{\pi}_i$  is obtained by 
\begin{align}
\tilde{q}_i = \text{MLP}(q_i), \quad
{\boldsymbol{\pi}}_i = \text{Sigmoid}(\tilde{q}_i C^\top), 
\label{eq:memberships}
\end{align}
% \begin{align}
% \tilde{q}_i &= \text{MLP}_{\pi}(q_i), \quad \hat{q}_j = \text{MLP}_{\rho}(q_j) \\ 
% {\boldsymbol{\pi}}_i &= \sigma(\tilde{q}_i C^\top) \in [0,1]^K, {\boldsymbol{\rho}}_j = \sigma(\hat{q}_j C^\top) \in [0,1]^K
% \label{eq:memberships}
% \end{align}
where
$\text{MLP}$ is a two-layer MLPs with ReLU activations that project queries to the community space.

\textbf{(4) Node-Specific Degree Correction:}
To account for  heterogeneous degree  distribution of tokens, we introduce a node-specific degree parameter for each token,
\[
% \gamma_i=\sigma(W_\gamma x_i),\quad
\theta_i=\text{Sigmoid}(W_\theta q_i)
\]
where  $W_\theta\in\mathbb R^{1\times d}$ are learned parameters.

\textbf{(5) Constructing the Connecting Probability.}
Combining all learned components, each entry of the community structure matrix is:
\begin{equation}
  p_{ij}
  \;=\;
  % \gamma_i\gamma_j\,
    \theta_i\theta_j\,
    {\boldsymbol{\pi}}_i^{\!\top} B\,{\boldsymbol{\pi}}_j
  \;
  \label{eq:cp_dcmmsbm_kernel_transformer}
\end{equation}

\noindent With $P$ constructed,  the DCMMM multi-head self-attention ($\mathrm{MSA}_{DCMM}$) is:
$$
\mathrm{MSA}_{DCMM}(Q, K, V, P) = \mathrm{Concat}(\mathrm{head}_1, \ldots, \mathrm{head}_h) W^O,
$$
where each attention head is defined as
$\mathrm{head}_i = \mathrm{Attn}_{\mathrm{DCMM}}(Q W_i^Q, K W_i^K, V W_i^V, P)$.
The learnable parameter matrices $W_{i}^{Q}$, $W_{i}^{K}$, $W_{i}^{V}$ are the projections in each subspace (head), and $W^{O}$ is the projection for all subspaces. Multi-head attention helps the model to jointly aggregate information from different representation subspaces at various positions.
% In this work, we apply the learned matrix $P$ to each representation subspace.
% The mask is sampled as
% \(M_{ij}\sim\operatorname{Bernoulli}(\pi_{ij})\).

% After obtaining $M_{ij}$, the following steps are the same as SBM-Transformer \citep{cho2022transformers}. 
% We extend the current structure to CP-DCMM (with configuration‑type periphery structure) in the future. 

\paragraph{Learning Objective and Loss Function.}
To enhance interpretability and encourage the model to learn distinct and meaningful community assignments for each token, we add an entropy regularization term on the soft community membership vectors.  Specifically, the entropy loss $\mathcal{L}_{\text{entro}}$ penalizes uniform (high-entropy) memberships and promotes sharper, more confident assignments, where each token clearly belongs to a specific set of communities.
Mathematically, the entropy of community assignment loss is:
\begin{equation}
    \mathcal{L}_{\text{entro}}= - \frac{1}{n}\left[ \sum_{i=1}^{n} \sum_{j=1}^{L} \pi_{ij} \log \left( \max(\pi_{ij}, \epsilon) \right) \right],
\end{equation}
where  $\epsilon$ is a small constant to prevent taking the log of zero.
The total loss is:
\begin{equation}
    \mathcal{L}_{\text{total}} = \mathcal{L}_{\text{task}} + \alpha\, \mathcal{L}_{\text{entro}},
\end{equation}
where $\mathcal{L}_{\text{task}}$ is the primary task loss (e.g., cross-entropy for medical image classification), and $\alpha$ is a hyperparameter balancing the entropy regularization. All model parameters, including cluster embeddings, membership projections, degree corrections, and transformer weights, are optimized jointly in an end-to-end manner using stochastic gradient descent.

\section{Experiments}

\subsection{Experiments Setup}

\textbf{Dataset.} 
We apply our proposed  DCMM-Transformer for the image classification task to evaluate its performance. 
We consider the following five diverse and representative medical imaging datasets: ChestXray, SIIM-ACR, INbreast, ADNI, and EyeDisease, each addressing different anatomical structures and imaging modalities. 

% \begin{itemize}
% (1) \textbf{ChestXray}: The ChestX-ray dataset~\cite{wang2017chestx}, encompasses an extensive collection of 112,120 chest X-ray images derived from 30,805 distinct patients, each annotated with disease labels.  Predominantly, this dataset focuses on two critical categories: pneumonia and normal states. Rewrite concisely.
(1) \textbf{ChestXray}: The ChestX-ray dataset~\cite{wang2017chestx} contains large-scale chest X-ray images. Here we focus on images of two major categories (5,856 in total):  pneumonia and normal. % (72.97\%)  (27.03\%)
% (2)  \textbf{SIIM-ACR}: The SIIM-ACR dataset~\cite{stephens2019acr}, is facilitated by the Society for Imaging Informatics in Medicine (SIIM) in collaboration with the American College of Radiology (ACR). The dataset is composed of 12,047 radiographic images, among which 2,669 are annotated to delineate the presence or absence of pneumothorax, categorically differentiated into two distinct classes: normal lung function and collapsed lung, the latter indicative of pneumothorax.
(2) \textbf{SIIM-ACR}: The SIIM-ACR dataset~\cite{stephens2019acr} includes 1,250 labeled radiographic images, classified as either normal lung or collapsed lung. %(78.56\%) 21.44\%
% (3)   \textbf{INbreast}: The INbreast database~\cite{moreira2012inbreast} is a mammographic database, with X-ray images acquired at a Breast Center, located in a hospital in Portugal. INbreast has a total of 115 cases (410 images) of which 90 cases are from women with both breasts (4 images per case) and 25 cases are from mastectomy patients (2 images per case).
(3)   \textbf{INbreast}: The INbreast database~\cite{moreira2012inbreast} consists of 6,154 mammography images collected at a breast center in Porto, Portugal. Each image is annotated with one of three classification labels: 
malignant, benign, and normal. % (33.13\%) (34.43\%) (32.44\%)
 % It includes 90 cases with images of both breasts (4 images per case) and 25 cases from mastectomy patients (2 images per case).
(4)   \textbf{ADNI}: 
% We use the structural connectivity (SC) constructed from MRI images, and the pre-processing follows the procedures established in existing works~\cite{Zhang2023Representative}. We use the public Alzheimer's Disease Neuroimaging Initiative (ADNI) dataset, which is the most widely used Alzheimer's disease dataset~\cite{yu2023supervised}. Through SC construction, 282 cognitively normal (CN) and 149 mild cognitive impairment (MCI) subjects are available. 
We use structural connectivity (SC) matrices derived from MRI scans in the publicly available Alzheimer's Disease Neuroimaging Initiative (ADNI) dataset~\cite{yu2023supervised}. The final dataset includes 282 cognitively normal (CN) subjects and 149 mild cognitive impairment (MCI) subjects. %following established preprocessing protocols~\cite{Zhang2023Representative}  65.4\%  34.6\% based on classification labels and their proportions.
(5)   \textbf{EyeDisease}:  The EyeDisease dataset~\citep{Zanlorensi22Ocular} contains approximately 4,000 high-resolution retinal (fundus) images, aggregated from multiple public sources. Each image is labeled with one of four categories: Normal, Diabetic Retinopathy (DR), Cataract, and Glaucoma, with roughly 1,000 images per class.  %(25\%)
% \end{itemize}

\textbf{Baselines.}  We compare with six state-of-the-art ViT baselines: ViT \citep{dosovitskiy2021image}, DeiT  \citep{pmlr-v139-touvron21a}, Swin Transformer  \citep{liu2021swin}, MaxViT  \citep{tu2022maxvit}, ConViT  \citep{d2021convit}, and SBM-Transformer  \citep{cho2022transformers}.
These models cover a broad spectrum of design philosophies, including standard transformers (ViT, DeiT), hierarchical and locality-aware attention (Swin, MaxViT, ConViT), and explicit community structure modeling (SBM-Transformer). This selection ensures that our evaluation is comprehensive.

\textbf{Training Details.} 
Since the dataset sizes are medium to small in scale, we build our DCMM-Transformer on ViT-Small (ViT-S) and initialize it with pre-trained weights~\cite{dosovitskiy2021image}. For fair comparison, all baseline models also use the small variant with pretrained weights. We use the AdamW optimizer, with a training batch size of 16 and a total of 100 training epochs. The learning rate is linearly increased from 0 to 0.0005, and subsequently follows a cosine decay schedule. Other training hyperparameters (such as dropout rate) are kept consistent across all experiments and methods for fair comparison. Our $\our$ introduces three specific hyperparameters: the entropy regularization weight $\alpha$, the community bias strength $\lambda$, and the number of communities $L$. After the grid search, we set the hyperparameters to $\alpha=0.1$, $\lambda=10$, and $K=100$. These values are used as defaults for $\our$, but all three hyperparameters can be further optimized using cross-validation. All experiments are conducted on NVIDIA H100 GPUs.
%is conducted using mini-batch stochastic gradient descent (SGD) with a momentum of 0.9 and a.  \ref{AS_Lambda}, and \ref{AS_Alpha}. We performed a sensitivity analysis in Figure~\ref{AS_NumCommunity}, finding that the optimal performance is achieved

% All other neural network hyperparameters (e.g., dropout rate, learning rate) are selected using cross-validation. 

\begin{figure}[!htb]
\centering
\includegraphics[width=1.0\columnwidth]{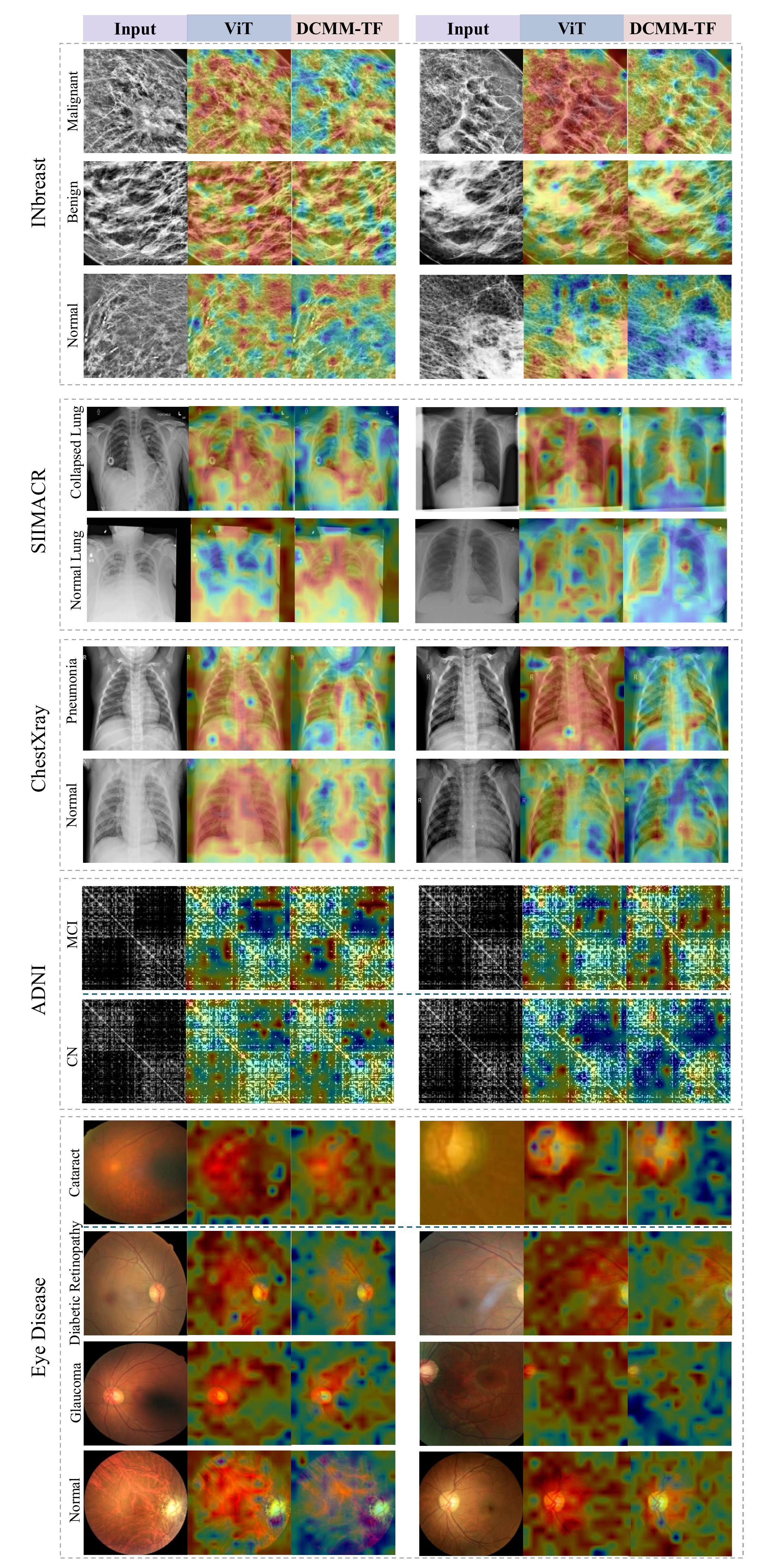}
\vspace{-10pt}
\caption{Visualization of attention maps from a standard Transformer and the proposed DCMM-Transformer (abbreviated as DCMM-TF in the figure) on the five datasets. The left column shows the original input images, the middle column displays attention maps from the standard Transformer, and the right column presents attention maps from $\our$. For each category, two random subjects are displayed.}
\label{DCMMAttXray}
\end{figure}

\subsection{Exepriment Results}

Table~\ref{tab:method_comparison} presents classification accuracies on five datasets. 
% comparing our DCMM-Transformer with state-of-the-art Vision Transformer baselines: ViT, DeiT, Swin Transformer, MaxViT, ConViT, and SBM-Transformer.
The DCMM-Transformer consistently outperforms all baselines across all datasets, achieving an average accuracy of $87.3\%$, surpassing the next-best method (MaxViT, $85.8\%$) by a significant margin. Notably, DCMM demonstrates the strongest gains on the ADNI dataset ($74.1\%$), highlighting its capacity to capture subtle structural variations in brain connectivity graphs, a challenging task where standard vision transformers underperform. The model also achieves state-of-the-art performance on the ChestXray ($96.0\%$) and SIIM-ACR ($88.2\%$) datasets, showcasing its effectiveness on 2D radiographic images with diverse pathological characteristics. These results validate the ability of the proposed degree-corrected mixed-membership attention mechanism to enhance representation learning in medical image classification tasks.

It is worth noting that the SBM-Transformer underperforms relative to standard baselines in this setting. This is likely because SBM-Transformer was originally designed to promote sparsity and reduce computational cost in NLP tasks, rather than to model the nuanced spatial and anatomical structures present in medical images. In contrast, the DCMM-Transformer is explicitly designed to exploit the spatial consistency and community-like organization of medical image patches. Moreover, in clinical practice, model performance and interpretability are typically prioritized over computational cost, reinforcing the practical value of our approach. These results validate the effectiveness of the proposed degree-corrected mixed-membership attention mechanism in enhancing representation learning for medical image classification.
%interpretability

\begin{table*}[!htb]
\centering
\caption{Ablation study on the effect of key components in DCMM-Transformer across five medical imaging datasets. \textbf{w/o $\mathcal{L}_{\text{entro}} $ }: without the structural constraint loss; \textbf{w/o DC}: without the degree correction module; \textbf{w/o MM}: without the mixed-membership module.}
\vspace{-5pt}
\begin{tabular}{ccccccc}
\hline
Method           & ChestXray & SIIMARC & INbreast & ADNI & EyeDisease  &Average \\ \hline
DCMM-Transformer & 96.0      & 88.2    & 85.1     & 74.1 & 93.3    & 87.3   \\
w/o $\mathcal{L}_{\text{entro}} $     & 95.2      & 84.2    & 83.8     & 66.7 & 92.3    & 84.4  \\
w/o DC           & 94.9      & 86.9    & 83.8     & 66.7 & 92.9    & 85.0   \\
w/o MM           & 95.6      & 87.5    & 81.8     & 65.4 & 93.0     & 84.7  \\ \hline
\end{tabular}
\label{tab:AScomponent}
\end{table*}

\begin{figure*}[!ht]
\centering
\includegraphics[width=2.1\columnwidth]{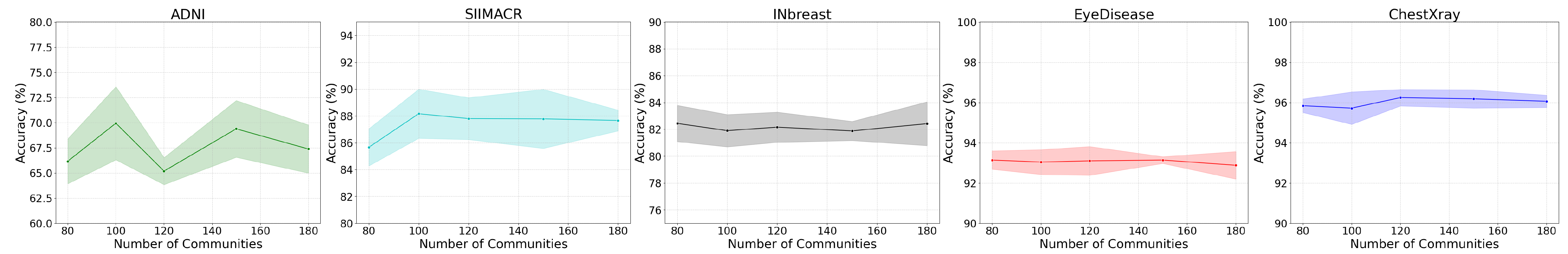}
\vspace{-10pt}
\caption{Influence of the number of communities on classification performance across five medical image datasets. The mean classification accuracy and standard deviation over three independent runs for each setting are reported. }
\label{AS_NumCommunity}
\end{figure*}

% As illustrated in Figure~\ref{AS_Lambda}, DCMM Attention exhibits robustness to a wide range of bias strengths, with moderate values generally leading to improved or stable performance.   These findings suggest that while DCMM bias benefits medical image feature learning. Moderate bias strength effectively integrates structural priors without overwhelming learned attention, contributing to both accuracy and interpretability. 

\subsection{Interpretability
}
Beyond improving classification accuracy, the proposed DCMM-Transformer mechanism significantly improves the interpretability of ViTs on medical imaging tasks. To  assess interpretability, we visualize the attention maps generated by our DCMM-Transformer as well as those from standard attention mechanisms across these datasets. Figure~\ref{DCMMAttXray} illustrates representative results on the five medical datasets. Each row in the figure corresponds to a specific diagnostic category (such as malignant, benign, or normal for INbreast). For each category, two representative image samples are shown. In each sample, the first column displays the original input image, followed by the attention heatmap generated by a standard ViT, and finally the attention map produced by our DCMM-Transformer. $\our$  demonstrates clinically superior interpretability, as evidenced by the following observation examples. 
%Additional visualizations for the EyeDisease and ADNI datasets are provided in  Figure \ref{DCMMAttADNIEye} in the Appendix. 
 
 (1) INbreast Mammography: In malignant cases, the $\our$ attention maps are tightly concentrated around the dense, spiculated masses, which are classic radiological indicators of malignancy, providing precise localization with minimal spillover into adjacent, non-pathological tissue. In contrast, ViT’s attention is more diffuse, often extending across wide areas of the breast, including normal fibroglandular tissue and even the pectoral muscle, which reduces clinical clarity.
For both benign and normal INbreast cases, $\our$ continues to produce attention maps that coherently follow the structure of the glandular tissue, showing respect for natural anatomical boundaries. 

 (2)   SIIM-ACR Chest X-ray: For collapsed lung (pneumothorax) cases, the DCMM-Transformer directs attention to the lateral and apical lung zones, precisely where a radiologist would search for the pleural line and absence of peripheral lung markings, key signs of pneumothorax. In particular, the model’s attention aligns well with the visible pleural edge and adjacent lucent area, which are crucial for accurate diagnosis. The ViT, by contrast, tends to distribute its attention as a vague cloud across the lung, offering little specific guidance on where to look for pathology.
In normal chest X-rays, $\our$ generally restricts focus within the lung fields and avoids irrelevant areas like the diaphragm and chest wall.

%\sw{How do learned communities align with medical domain knowledge? }

% This improved interpretability is due to DCMM’s structured bias, which encodes soft community membership and degree information directly into attention scores. These inductive biases promote learning of spatially coherent and semantically aligned token interactions, resulting in attention distributions that are not only discriminative but also clinically interpretable. Consequently, DCMM-Transformer not only enhances prediction accuracy but also improves model transparency, making it more suitable for deployment in real-world medical applications.

\subsection{Ablation and Sensitivity Analysis}

%\sw{computational cost analysis}

\textbf{Ablation. }
We conducted comprehensive ablation studies to evaluate the impact of the key components of $\our$, and the entropy penalty term on model performance across five medical imaging datasets. Each component was removed individually to isolate its effect, and the results are summarized in Table~\ref{tab:AScomponent}. Specifically, omitting the entropy loss leads to the largest average performance drop (from $87.3\%$ to $84.4\%$), highlighting its critical role in encouraging confident, interpretable community assignments. Eliminating degree correction or mixed-membership also decreases accuracy (to $85.0\%$ and $84.7\%$ average, respectively), underscoring the value of modeling both node connectivity heterogeneity and overlapping community structure.

\textbf{Sensitivity.} Our sensitivity analysis shows that DCMM-Transformer is remarkably stable across a broad range of hyperparameter settings. Among them, the number of communities is particularly important, as it directly and indirectly influences the DCMM bias. We therefore further examine the model’s performance under different numbers of communities. As seen in Figure~\ref{AS_NumCommunity}, varying $L$ from 80 to 180 yields stable classification accuracy, especially on INbreast, EyeDisease, and ChestXray. This means the model doesn’t require precise tuning of this parameter; it works well over a broad range, which is practical in real medical applications. 

\section{Conclusion}
In this work, we proposed the DCMM-Transformer, a novel Vision Transformer architecture that integrates the DCMM community structure into the attention mechanism. 
By learning soft group memberships and degree scalars directly from the query representations, and integrating them into the attention mechanism via an additive bias, DCMM-Transformer enables the model to focus on semantically coherent and clinically relevant regions within medical images. 
This approach not only enhances classification accuracy but also improves interpretability. Extensive experiments across diverse medical imaging datasets demonstrate that DCMM-Transformer consistently outperforms strong transformer baselines, while ablation studies validate the individual contributions of the degree correction, mixed-membership, and entropy regularization modules.

\section{Acknowledgments}
This work was partially supported by the U.S. National Science Foundation (NSF) under grants  DMS-1903226, DMS-1925066, DMS-2124493, DMS-2311297, DMS-2319279, DMS-2318809, and by the U.S. National Institutes of Health (NIH) under grant R01GM152814. 
Any opinions, findings, conclusions, or recommendations expressed in this material are those of the authors and do not necessarily reflect the views of the NSF, or NIH.

\newpage
\bibliography{CameraReady/LaTeX/aaai2026}

\iffalse
\clearpage
\newpage
\appendix
\fi

% \onecolumn % check this
%\input{CameraReady/LaTeX/appendix}

\clearpage

\end{document}